\documentclass[10pt,twocolumn,letterpaper]{article}
\usepackage{authblk}
\usepackage{cvpr}              

\usepackage{graphicx}
\usepackage{amsmath}
\usepackage{amssymb}
\usepackage{booktabs}
\usepackage{array}
\usepackage{algorithm}
\usepackage{algorithmic}
\usepackage{multirow}

\usepackage[pagebackref,breaklinks,colorlinks]{hyperref}

\usepackage[capitalize]{cleveref}
\crefname{section}{Sec.}{Secs.}
\Crefname{section}{Section}{Sections}
\Crefname{table}{Table}{Tables}
\crefname{table}{Tab.}{Tabs.}

\begin{document}


\title{Text2Grasp: Grasp synthesis by text prompts of object grasping parts.}

\author[1]{Xiaoyun Chang \thanks{cgsmalcloud83@mail.dlut.edu.cn}}
\author[1]{Yi Sun \thanks{ Corresponding author, lslwf@dlut.edu.cn}}
\affil[1]{Dalian University of Technology}

\maketitle

\begin{abstract}
The hand plays a pivotal role in human ability to grasp and manipulate objects and controllable grasp synthesis is the key for successfully performing downstream tasks. Existing methods that use human intention or task-level language as control signals for grasping inherently face ambiguity. To address this challenge, we propose a grasp synthesis method guided by text prompts of object grasping parts, Text2Grasp, which provides more precise control. Specifically, we present a two-stage method that includes a text-guided  diffusion model TextGraspDiff to first generate a coarse grasp pose, then apply a hand-object contact optimization process to ensure both plausibility and diversity.  Furthermore, by leveraging Large Language Model, our method facilitates grasp synthesis guided by task-level and personalized text descriptions without additional manual annotations. Extensive experiments demonstrate that our method achieves not only accurate part-level grasp control but also comparable performance in grasp quality.

\end{abstract}

\section{Introduction}
\label{sec:intro}

Modeling hand grasps have recently gained extensive attention due to its wide applications in  human-computer interaction \cite{pollard2005physically}, virtual reality \cite{holl2018efficient,wu2020hand}, and imitation learning in robotics \cite{hsiao2006imitation}. To predict plausible human-like grasp poses when given an object, many hand-object interaction datasets \cite{hasson2019learning, corona2020ganhand,hampali2020honnotate,chao2021dexycb} has been built to promote research \cite{taheri2020grab,li2022contact2grasp,liu2023contactgen} on learning human experience in recent years. However, these works concentrate on stable grasps that are not suitable for task-oriented grasps. Different tasks necessitate specific types of grasps. For instance, in a cutting task, people typically grasp a knife by its handle rather than the blade. Similarly, when handing over a knife, it is safer for the deliverer to hold the blade, minimizing the risk of injury to the receiver. Consequently, controllable grasp synthesis is of paramount important.

\begin{figure}[t]
\centering
\includegraphics[width=0.48\textwidth]{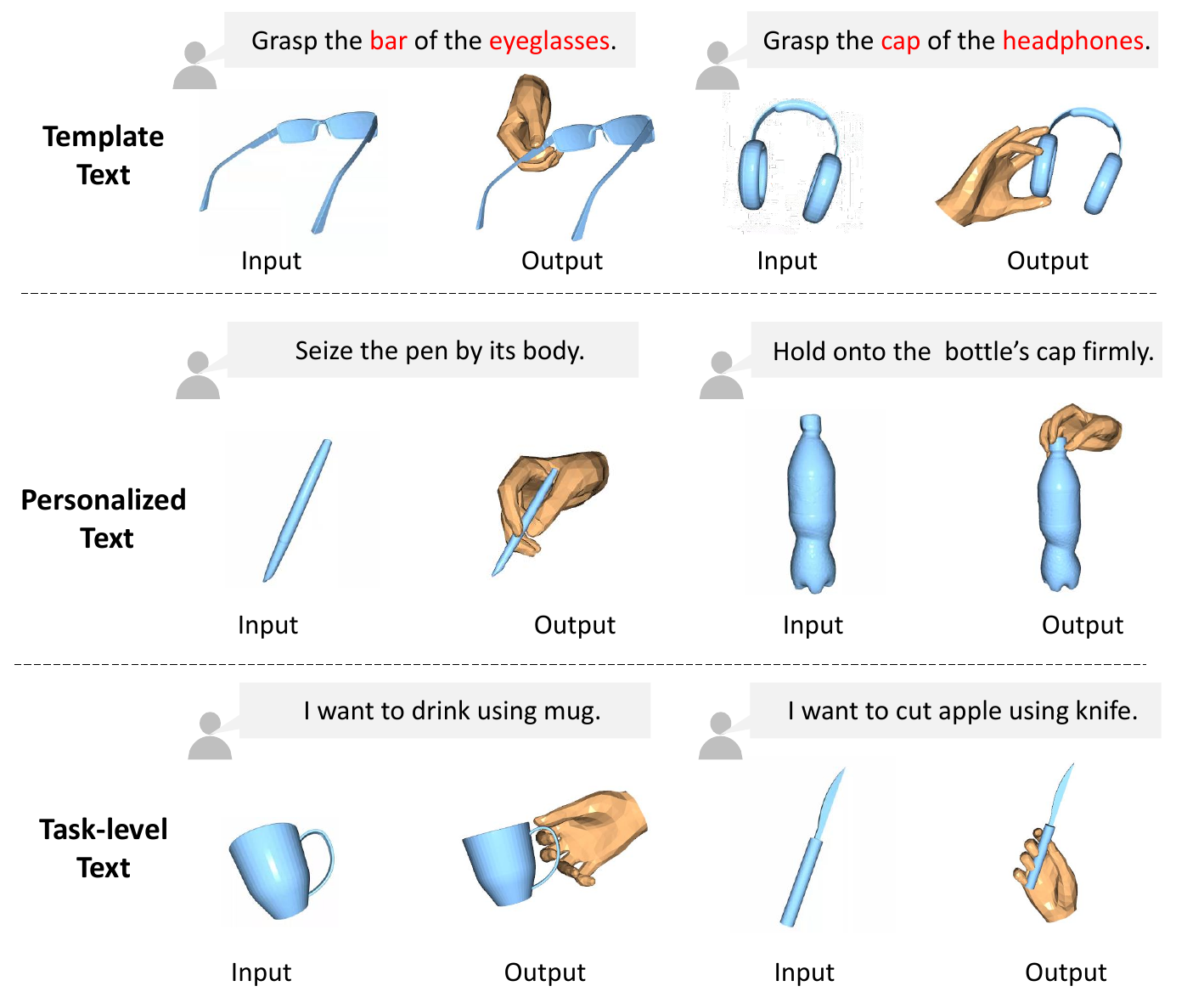}
\caption{Given an object, Text2Grasp can generate specific hand grasps by interpreting various text inputs: a) Template text. b) Personalized text. c) Task-level text. }
\label{teaser}
\end{figure}

To facilitate controllable synthesis, many studies \cite{brahmbhatt2020contactpose,taheri2020grab,yang2022oakink,jian2023affordpose} introduce various datasets of grasp containing different numbers of human intentions, such as use, pass, twist and so on. Furthermore, \cite{yang2022oakink} and \cite{jian2023affordpose} translate these intentions into one-hot embeddings, combining them with object point cloud feature to achieve intention-guided grasp synthesis. Considering that language is more natural mode of interaction, some studies \cite{song2023learning,tang2023task,nguyen2023language} start to employ task-level text descriptions as inputs for predicting 6-Dof pose of parallel jaw gripper. However, utilizing fixed set of intentions or task-level text descriptions for grasping inherently faces ambiguity, primarily in two aspects: 1) Same intention but different grasps. For instance, "lifting a mug" may involve different grasp types of either grasping the handle or the body. 2) Different intentions but same grasps. For instance, "lifting" or" twisting" might have  same initial grasping pose to hold a bottle's neck. Such complexities increase the difficulty in annotating datasets and achieving model convergence. 

To overcome these limitations, we propose a grasp synthesis method abbreviated as Text2Grasp that is guided by text prompts of object grasping parts, rather than intentions or task descriptions that are  unable to explicitly indicate which part of the object to grasp. Text2Grasp takes an object and a predefined text template: \textit{Grasp the [Object Part] of the [Object Category]} as input, and generates a grasp pose targeting the specified part of the object for manipulation. This part-level guidance reduces uncertainty compared to intent-based or task-level guidance, facilitating better convergence of the  grasp generation network. Specifically, we present a two-stage method that includes a text-guided  diffusion model TextGraspDiff to first generate a coarse grasp pose, then apply a hand-object contact optimization process to ensure both plausibility and diversity. Unlike all-finger optimization approaches that prioritize maximum object-finger contact\cite{corona2020ganhand}—resulting in mainly closed-finger grasps—our optimization emphasizes optimizing contact between the  fingers and the  object part specified by text description. This strategy ensures physical realism, diversity in grasps, and alignment with text.

Furthermore, the template representation of text prompts for the object grasping parts also supports grasp synthesis guided by task-level and personalized text prompts, since LLM \cite{brown2020language} has been able to divide the task descriptions into several execution steps, including the parts of the object that should be grasped. Subsequently, our Text2Grasp can generate task-level grasps taking the inference results of LLM as input. Moreover, the utilization of LLM allows for the expansion of our designed text templates, enriching the training dataset for personalized text descriptions.

In summary, our contributions are as follows:

$\bullet$  We propose Text2Grasp, a grasp synthesis method guided by text prompts of object grasping parts, offering a more natural interaction and precise grasp control.

$\bullet$  We introduce a two-stage method that includes a text-guided  diffusion model TextGraspDiff to first generate a coarse grasp pose, then apply a hand-object contact optimization process to ensure both plausibility and diversity.

$\bullet$  By leveraging LLM, our method facilitates grasp synthesis guided by task-level  and personalized text descriptions  without additional manual annotations.

Extensive experiments on public datasets demonstrate that our method achieves not only accurate part-level grasp control but also comparable performance to state-of-the-art methods in terms of grasp quality.

\section{Related Work}\label{sec:2}

There has been a significant amount of research in the field of  grasp synthesis. Here, we focus on realistic human grasp synthesis and review the most relevant works. Based on whether the grasp generation is controllable, we categorize the synthesis algorithms into two types: Uncontrolled Grasp Synthesis and Controllable Grasp Synthesis.

\textbf{Uncontrolled Grasp Synthesis.} Uncontrolled grasp synthesis primarily aims to generate hand pose capable of stably grasp objects without considering subsequent tasks. A trend has emerged to develop deep learning solutions, driven by the introduction of large-scale datasets of hand-object interactions \cite{corona2020ganhand,hampali2020honnotate,hasson2019learning,taheri2020grab,brahmbhatt2020contactpose,yang2022oakink,jian2023affordpose}. These methods learn the latent distribution of hand-object contact information or hand parameters through generative models, including Generative Adversarial Network (GAN) \cite{goodfellow2020generative} and Conditional Variational Auto-Encoder (CVAE)\cite{sohn2015learning}. GanHand\cite{corona2020ganhand} initially predicts the optimal grasp type from a taxonomy of 33 classes, and then employs a discriminator and an optimization to get a refined grasp. Instead of predicting MANO\cite{romero2017embodied} parameters directly, ContactDB \cite{brahmbhatt2019contactdb} use thermal cameras to capture object contact maps that reflect the contact regions of an object post-grasping and utilizes GAN to learn their distribution, facilitating grasp synthesis. 

Comparing with GAN\cite{goodfellow2020generative}, CVAE\cite{sohn2015learning} are more popular in hand grasp synthesis because of its simple structure and one-step sampling procedure. GrabNet\cite{taheri2020grab} utilizes CVAE by conditioning on the Basis Point Set \cite{prokudin2019efficient} of objects and samples from the low-dimensional space mapped through CVAE to generate hand grasps. Additionally, it incorporates a neural network to refine the coarse pose. This approach is also followed by Oakink \cite{yang2022oakink} and AffordPose \cite{jian2023affordpose}. Grasp Field \cite{karunratanakul2020grasping} and HALO \cite{karunratanakul2021skeleton} learn an implicit grasping field using CVAE as the hand representation to produce high-fidelity hand surface. GraspTTA \cite{jiang2021hand} exploits the contact map introduced by ContactDB\cite{brahmbhatt2019contactdb} to refine the grasps generated by CVAE during reference. Contact2Grasp \cite{li2022contact2grasp} learns the distribution of contact map for grasps by CVAE and then maps the contact to grasps. Moreover, ContactGen\cite{liu2023contactgen} introduces a  three-component model to represent the contact of hand-object: the contact probability, the specific hand part making contact, and the orientation of the touch, and a sequential VAE is proposed to learn these aspects for grasp synthesis. Despite its simplicity and direct sampling process, CVAE often suffer from the posterior collapse\cite{yuan2020dlow,wang2021synthesizing,huang2023diffusion}. This leads to less diverse outputs, including simplistic samples like a slightly closed hand shape. To mitigate this problem, SceneDiffuser\cite{huang2023diffusion}, UGG \cite{lu2023ugg} and DexDiffuser\cite{weng2024dexdiffuser} employ a diffusion-based denoising process, ensuring diverse sample generation by gradually denoising, thus avoiding direct latent space mapping. 

These aforementioned methods are capable of generating stable grasps. However, these grasps might not be consistent with human manipulation habits, making them less appropriate for the tasks. Consequently, instead of solely relying on  object shape as input, we incorporate the text prompts of object grasping parts into diffusion model for controllable grasp synthesis. Moreover, in contrast to methods that utilize global optimization \cite{corona2020ganhand, taheri2020grab} to refine grasps, our work introduces an optimization based on finger perception and object part perception. This strategy not only ensures grasp stability  but also maintains diversity.

\textbf{Controllable Grasp Synthesis.} The capacity for controllable grasping is crucial as it represents the first step for manipulation. To facilitate controllable grasp synthesis, datasets \cite{brahmbhatt2020contactpose,taheri2020grab,yang2022oakink,jian2023affordpose}
 encompass a range of human intentions for dexterous hand grasping. ContactPose \cite{brahmbhatt2020contactpose} identifies two basic intentions: use and pass. Expanding on this, GrabNet\cite{taheri2020grab} introduces lifting and off-hand passing. OakInk\cite{yang2022oakink} goes further by incorporating intentions such as holding and receiving. AffordPose \cite{jian2023affordpose} elaborates on the use intention, creating hand-centric categories like twisting, pulling, handle grasping, among eight total intentions. Additionally, to generate intent-driven grasps, OakInk\cite{yang2022oakink} and AffordPose\cite{jian2023affordpose} translate these intentions into word embeddings, combining them with object point cloud features as the condition of CVAE to produce matching grasp pose. Considering that language is one of the most natural forms of human interaction, some studies employ task-level text descriptions as inputs for predicting grasps with parallel jaw gripper. These methods initially construct extensive datasets of grasps that include task-level text descriptions. Based on these datasets, \cite{song2023learning} and \cite{tang2023graspgpt} adopt a generate-then-select methodology. It involves initially generating a number of poses for parallel jaw gripper, followed by a selection process guided by task-level text descriptions. In contrast, \cite{tang2023task} and \cite{nguyen2023language} directly predict the position of gripper on the input RGB image or object point cloud based on task-level text description guidance. Comparing to the simple closing of a gripper, the human hand, with its higher degree of freedom, must not only ensure stable grasping but also maintain the rationality of itself and interaction, making grasp synthesis for it more challenging.

These methods, utilizing fixed set of intentions or task-level text descriptions  for grasping, inherently face ambiguity, especially when defining intentions or tasks for identical parts of an object, such as a mug's handle and body. To address this, we develop a grasp synthesis method guided by text prompts of object grasping parts. Compared to the ambiguity of intentions or task-level guidance, part-level guidance offers lower uncertainty, which facilitates the convergence of grasp synthesis networks. Furthermore, our method contrasts with those requiring manual labeling \cite{tang2023task,nguyen2023language,tang2023graspgpt,song2023learning}, by leveraging Large Language Model \cite{brown2020language} to facilitate grasp synthesis guided by task-level and personalized text descriptions without additional manual labels.

\begin{figure*}[ht]
\centering
\includegraphics[width=0.98\textwidth]{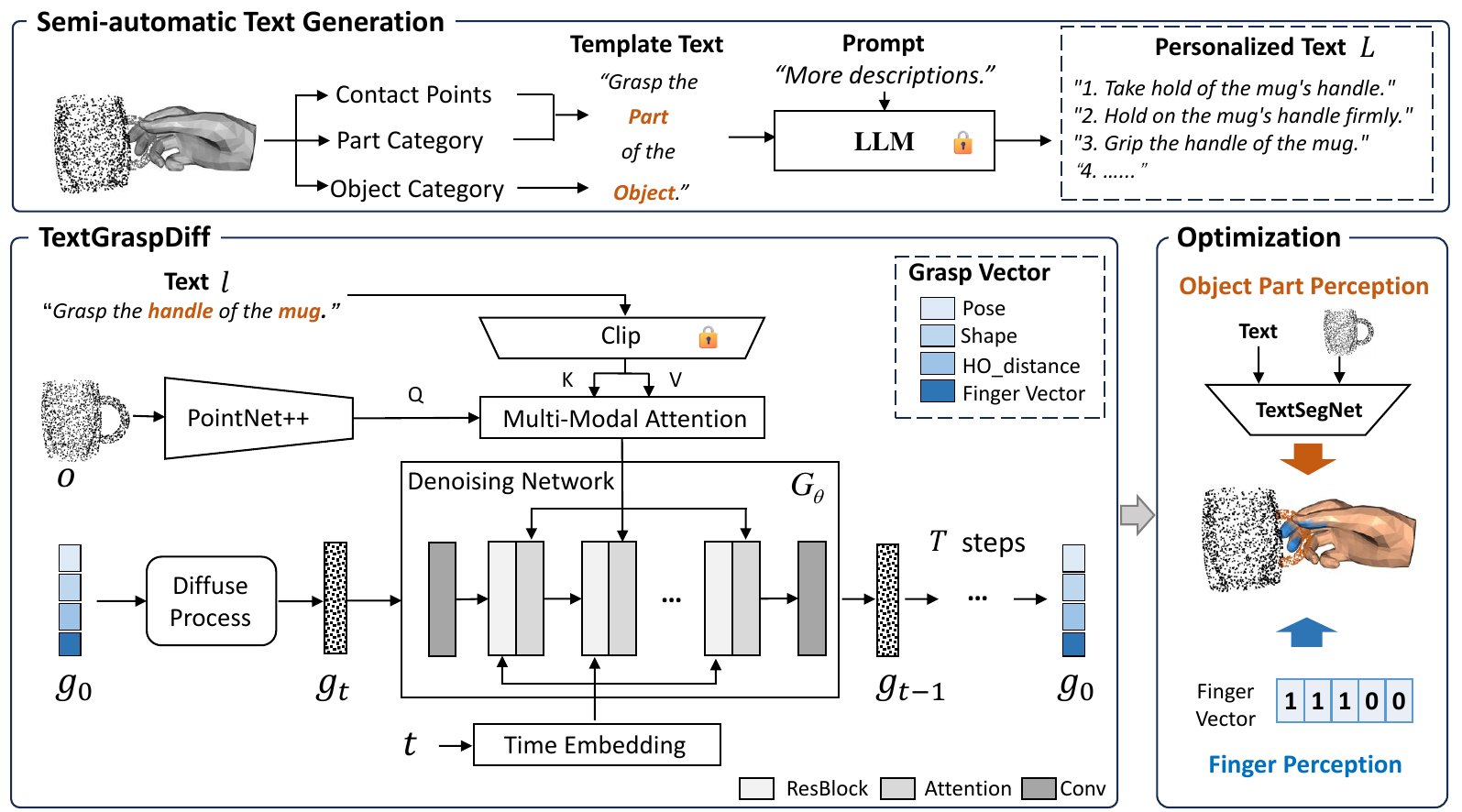}
\caption{\textbf{The Overview of Text2Grasp.}  We present a semi-automatic approach to generate both the template text and the personalized text prompts for each grasp in the datasets, which are used to train TexGraspDiff.
And given the  point cloud of object and  text description of object grasping parts, we introduce a two-stage method that includes a text-guided  diffusion model TextGraspDiff to first generate a coarse grasp pose, then apply a hand-object contact optimization process to ensure both plausibility and diversity. The final hand mesh can be obtained by MANO model\cite{romero2017embodied}.}
\label{fig:overview}
\end{figure*}

\section{Methods}\label{sec:3}

Our aim is to achieve controllable grasp synthesis when given an object's point cloud and a text prompt of object grasping part, ensuring the generated hand grasps stably hold the object while aligning with the input text. To this end, we introduce a two-stage method that includes a text-guided  diffusion model TextGraspDiff to first generate a coarse grasp pose, then apply a hand-object contact optimization process to ensure both plausibility and diversity. The overview of our method is illustrated in \cref{fig:overview}. 

In this section, we first present our semi-automated text generation method in \ref{sec:3.1}. We then detail the text-guided conditional diffusion model-TextGraspDiff in \ref{sec:3.2}, and the hand-object contact optimization in Section \ref{sec:3.3}. 

\subsection{Semi-automatic Text Generation for Grasp}\label{sec:3.1}

The key idea behind Text2Grasp is to leverage text prompts of object grasping parts to control grasp synthesis. Instead of relying on extensive manual annotations, which are extremely labor-intensive and time-consuming, we design a semi-automatic approach to generate text prompts for existing hand grasp dataset, as illustrated in \cref{fig:overview}. First, we predefined the text template, \textit{i.e.}, \textit{Grasp the [Object Part] of the [Object Category]}. The object category can be directly provided by existing datasets, while the object part corresponding to each grasp can be determined through computation. Specifically, given the point cloud of an object and the hand mesh grasping it, we first calculate the contact between the object and the hand, assigning a contact label to each point on the object. And  the \textit{``Object Part''} label for each grasp is determined by the object part  with the most contact points. Finally, we can generate a text template for each grasp in the datasets. 

Furthermore, we leverage Large Language Model \cite{brown2020language} with strong text comprehension capabilities to expand the template text, thereby generating more personalized text descriptions. For example, given the prompt \textit{``Please write [N] sentences with the same meanings as [template].''}, where $N$ is the
number of generated text descriptions, LLM can then infer a variety of plausible text descriptions to form our candidate text list $L$. During training, we randomly select one description from $L$ as a training label for each grasp. 

This semi-automatic text generation approach facilitates personalized text inputs, thereby enhancing the flexibility of grasp synthesis control. In addition, the representation of the text prompts for the object grasping parts gives our method the ability to achieve task-level grasp synthesis because Large Language Model \cite{brown2020language} can provide a description of the grasping action from a task description, such as grasping the mug's handle for drinking task. Thus we can accomplish task-level grasp synthesis without extra training.

\subsection{Text to Grasp via  Diffusion Model}\label{sec:3.2}

In this section, we introduce TextGraspDiff, a conditional diffusion model for grasp synthesis that is guided by text prompts of object grasping part. The overview of our method is illustrated in \cref{fig:overview}. Taking the object point cloud $o\in R^{N\times 3}$ and the part-level text prompts $l$, TextGraspDiff outputs a hand grasp vector $g\in R^{66}$. This grasp vector encompasses MANO \cite{romero2017embodied} model pose $g_{\theta}\in R^{48}$, shape $\quad g_{\beta}\in R^{10}$, the distance $g_{dis}\in R^3$  between object and hand centroids, and the finger vector $g_{f}\in R^5$, which indicates which fingers are being used for grasping. Adhering to the diffusion model outlined in \cite{ho2020denoising}, our method is comprised of both a forward process and a reverse process.

\textbf{Forward process.} Given a grasp vector $g_0$, sampling from the ground-truth data distribution, we add the infinitesimal Gaussian noise $\epsilon_t\sim N(0,\beta_tI)$ into $g_{0}$ and get a sequence of noised data  $\{g_i\}_{t=1}^T$  after $T$ step. $\beta_t$ adheres to a linear variance schedule. 
\begin{equation}
q(g_t\mid g_0)=\mathcal{N}(g_t;\sqrt{\bar{\alpha}_t}g_0,(1-\bar{\alpha}_t)I)
\label{eq:1}
\end{equation}
 where $\alpha_t=1-\beta_t$, $\overline{\alpha}_t=\prod_{s=1}^t\alpha_s$. After $T$ steps, if the amount of noise added is sufficiently large, then $g_T$  approximately converges to a standard Gaussian distribution.

\textbf{Reverse process.} The process reverses noise sampled from a Gaussian distribution back into a sample from the data distribution for a fixed timestep. In our work, with the grasp vector $g_{0}$ as the target for denoising process, and object point cloud $o$  and its text prompt $l$ of object grasping part as conditions, the conditional diffusion model leads to $p(g_{t-1}|g_{t},o,l)$. Following \cite{tevet2022human,qi2023diffdance}, we predict the grasp vector $g_{0}$ using a neural network $G_\mathrm{\theta}$. This process can be formalized as:
\begin{equation}
    p\big(g_{t-1}|\:g_t,o,l\big)=\mathcal{N}\bigg(g_{t-1};\mu_{\theta}(g_t,o,l,t),\widetilde{\beta}_tI\bigg)
    \label{eq:2}
\end{equation}
\begin{equation}
    \tilde{\mu}_\theta(g_t,o,l)=\frac{\sqrt{\bar{\alpha}_{t-1}}\beta_t}{1-\overline{\alpha}_t}G_\theta(g_t,o,l,t)+\frac{\sqrt{\alpha_t}(1-\overline{\alpha}_{t-1})}{1-\overline{\alpha}_t}g_t
    \label{eq:3}
\end{equation}

The detailed structure of the denoising network $G_\mathrm{\theta}$ is shown in \cref{fig:overview}, we employ a Transformer\cite{vaswani2017attention} as the denoising network's  backbone, which has demonstrated promising results in human motion synthesis\cite{tevet2022human,qi2023diffdance} and robotic hand grasp synthesis\cite{huang2023diffusion,lu2023ugg}. For multi-condition inputs, including point clouds and text, we initially employ the PointNet++ \cite{qi2017pointnet++} and the pretrained CLIP \cite{radford2021learning} model as respective encoders to extract the point cloud feature and text feature. Instead of simply adding these multi-modal features, we design a Multi-Modal Attention Module based on Transformer\cite{vaswani2017attention} for effective fusion, leveraging point cloud feature $f_p$  as the query and text feature  $f_l$  as the key and the value. This fusion mechanism enables us to achieve more precise control over grasp locations. Following \cite{huang2023diffusion}, we incorporate a timestep-residual block and cross-attention for input noise embedding feature-condition fusion to ensure the network is effectively guided by step  $t$ and the condition $c$. Finally, the grasp vector $g_0$ can be predicted by final output layer. The loss function of the network $G_\mathrm{\theta}$  is:
\begin{equation}
    \mathcal{L}=E_{g_0\sim q(g_0|o,l),t\sim[1,T]}[\left\|g_0-G_\theta(g_t,t,o,l)\right\|_2^2]
    \label{eq:4}
\end{equation}

After completing the training of the denoising network $G_\mathrm{\theta}$ , when given a new object's point cloud and  text description as conditions, we first sample random noise from a Gaussian distribution, then apply the denoising network $G_\mathrm{\theta}$  and \cref{eq:2}  and \cref{eq:3} over $T$ steps, and finally we obtain the grasp vector matching the object's part-level text description. The grasp hand mesh is then generated by applying the final grasp vector to the MANO model\cite{romero2017embodied}.

\subsection{Text-guided Contact Optimization}\label{sec:3.3}

To produce physically more plausible grasps, many works \cite{corona2020ganhand,jiang2021hand,hasson2019learning,li2022contact2grasp} introduce a refinement stage to enhance contact and minimize penetration. Their main focus is on stable grasping by aligning hand with the closest object surface points, but these points may not match the text-described object parts, potentially leading to inaccurate grasp locations. Therefore, we propose a text-guided contact optimization method based on finger perception and  text-guided object part perception. It guides specific fingers toward the object part described by text description, further enhancing grasp stability, diversity, and grasp part accuracy.

\begin{figure}[t]
\centering
\includegraphics[width=0.48\textwidth]{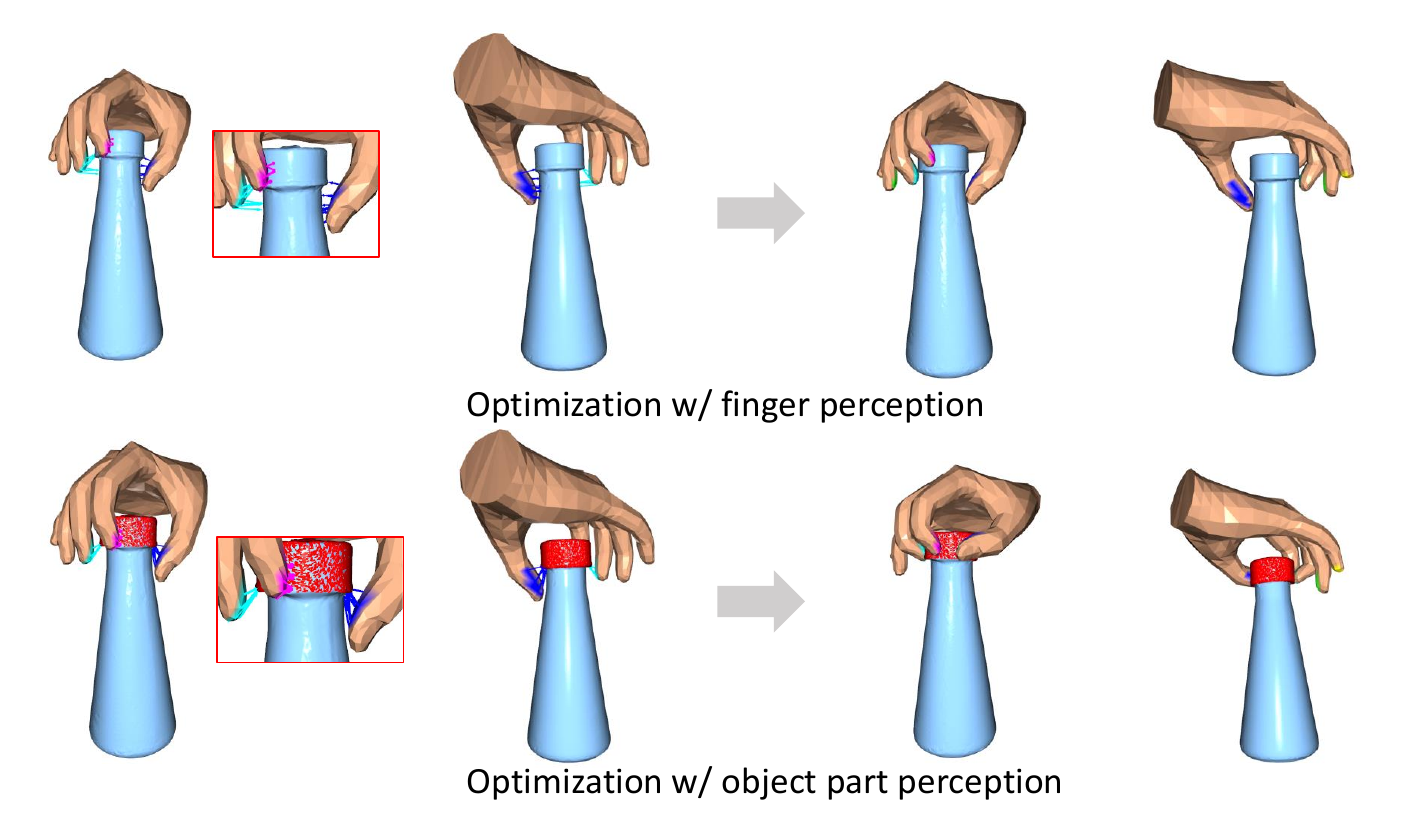}
\caption{\textbf{The Contact optimization.} The contact optimization consists of finger perception and  object part perception. The finger perception optimization directs the particular fingers used for grasping towards object and the object part optimization guided fingers toward the object part specified by text.}
\label{fig:method_refine}
\end{figure}

\textbf{Hand finger perception.} Rather than minimizing the distance between all prior hand contact vertices often utilized for grasping and object points, we specifically optimize the distance between the object and the particular fingers used for grasping. We utilize a five-dimensional finger  vector to define which fingers are used in grasping the object. For instance, as shown in \cref{fig:method_refine}, if the grasp involves using the thumb, index, and middle finger, then the finger  vector $g_{f}$ is [1,1,1,0,0]. Following human habits, this vector is generated alongside the grasp. During optimization, we minimize the distance only between the object and those fingers indicated by a 1 in the finger vector, avoiding the issue of all fingers contacting the object. The loss for the finger perception optimization is formulated as:
\begin{equation}
    H_c = \bigcup_{i=1}^{5} \{ C_i \mid g_{f}^i = 1 \}
    \label{eq:5}
\end{equation}
\begin{equation}
    \mathcal{L}_{_{hc}}(H_c,O)=\frac1{\mid H_c\mid}\sum_{h\in H_c}\min_k\left\|h,O_k\right\|
    \label{eq:6}
\end{equation}
where $C_i$  represents the set of hand vertices which belongs to the $i$th fingertips, statistics by \cite{hasson2019learning}. And $H_c$ denotes the set of points for all fingers making contact.

\textbf{Text-guided object part perception.} As shown in \cref{fig:method_refine} we minimize the distance between hand contact points and object part specified by the text input to guide hand fingers to grasp the correct object part. Specifically, using a pretrained text-guided segmentation network TextSegNet, we first segment the input object point cloud into targeted $O^{c}$ and non-targeted parts $O^{nc}$ based on the input text prompts of object grasping part. During the optimization, we assign higher weights to the targeted part, directing the hand contact points toward it to enhance the accuracy of the grasp part. The hand-object contact loss is formulated as follows:
\begin{equation}
    \mathcal{L}_c(H_c,O)=\lambda_1\mathcal{L}_{nc}(H_c,O^c)+\lambda_2\mathcal{L}_{nc}(H_c,O^{nc})
    \label{eq:7}
\end{equation}
where $\lambda_1$  and $\lambda_2$  are hyperparameters. $\quad O^c=\{p_i\in O|\mathrm{~}F_{\mathrm{seg}}(p_i,l)=1\}$ and $O^{nc}=\{p_i\in O|\mathrm{~}F_{_{\gcd}}(p_i,l)=0\}$ respectively represent the targeted and non-targeted parts. $F_{\mathrm{seg}}$  is the pretrained text-guided segmentation network TextSegNet. We also use PointNet++ \cite{qi2017pointnet++} and CLIP\cite{radford2021learning} for point cloud and text encoding, followed by a multi-layer fully connected network to output segmentation labels. The training loss utilizes negative log-likelihood loss. 

\textbf{Others.} Following \cite{corona2020ganhand,hasson2019learning}, we minimize the object points that are inside the hand distance to their closest hand surface points to penalize hand and object interpenetration by $\mathrm{L}_{ptr}$. Furthermore, following \cite{zhu2023toward}, we incorporate a joint angle limitation loss $\mathrm{L}_{angle}$ and a self-collision loss $\mathrm{L}_{self}$ for the hand to ensure the plausibility of the grasping hand pose. 

Ultimately, our overall optimization objective is formulated as follows:
\begin{equation}
\min_{g_\theta,g_\beta,g_{dis}}\lambda_c\mathcal{L}_c+\lambda_{ptr}\mathcal{L}_{ptr}+\lambda_{angle} \mathcal{L}_{angle} + \lambda_{self}\mathcal{L}_{self}
    \label{eq:8}
\end{equation}
where $\lambda_c$ , $\lambda_{ptr}$ , $\lambda_{angle}$ and $\lambda_{self}$ is a hyper-parameter. We utilize this objective function to optimize the network-predicted MANO model pose $g_\theta$ , shape $g_\beta$, the distance $g_{dis}$  between object and hand centroids, aiming to further enhance the quality of generated grasps and the part accuracy of grasp .

\begin{table*}[htbp]
\centering
\begin{tabular}{cm{2.5cm}ccccccc}
\toprule
\rule[-1ex]{0pt}{4ex}\multirow{2}{*}{\textbf{Dataset}} & \multirow{2}{*}{\textbf{Methods}} & \multicolumn{2}{c}{\textbf{Penetration}} &  \multirow{2}{*}{\begin{tabular}[c]{@{}c@{}}\textbf{Simulation} \\ \textbf{Displacement} \\ \textbf{Mean ± Var$\downarrow$}\end{tabular}}  & \multicolumn{2}{c}{\textbf{Diversity}} &  \multirow{2}{*}{\begin{tabular}[c]{@{}c@{}}\textbf{Part} \\ \textbf{Accuracy}$\uparrow$\end{tabular}}    \\ 
 \cmidrule{3-4} \cmidrule{6-7} \rule[-1ex]{0pt}{4ex}
                                 &             & \textbf{Depth$\downarrow$} & \textbf{Volume$\downarrow$} &  & \textbf{Entropy}$\uparrow$ & \textbf{Cluster Size}$\uparrow$    \\ 
\midrule
\multirow{4}{*}{\textbf{OakInk}} 
&  Test$_{GT}$  & 0.11  &  0.65  &  1.80 ± 2.04 & 2.91 & 4.11 &100.00 \\ 
&  GrabNet\cite{taheri2020grab} & 0.48 &  2.97 &  2.84 ± 2.81   & \textbf{2.95}   & 2.57  &   - \\ 
&  Ours$_{template}$ & \textbf{0.40}& 1.89& \textbf{2.49 ± 2.51} & 2.92 & 4.70  & \textbf{87.76}\\ 
&   Ours$_{personalized}$ & 0.41 & \textbf{1.73} & 2.49 ± 2.57 &2.92 &\textbf{4.74} &82.32  \\ 
\midrule
\multirow{3}{*}{\textbf{AffordPose}} 
& GrabNet \cite{taheri2020grab} & \textbf{0.54} & \textbf{3.77} & 3.09 ± 2.74 &\textbf{2.94} & 2.52 & -   \\ 
& Ours$_{template}$ & 0.66 & 5.05 & \textbf{2.93 ± 2.67} & 2.90 & \textbf{4.88} & \textbf{78.53}   \\  
&  Ours$_{personalized}$  & 0.59 & 3.84 & 3.00 ± 2.86&2.87&4.79&73.83  \\ 
\bottomrule
\end{tabular}

\caption{The quantitative results on the OakInk\cite{yang2022oakink} dataset and the AffordPose \cite{jian2023affordpose} dataset.Test$_{GT}$ means the grouth-truth grasps on the OakInk Test datasets.  Ours$_{template}$ and Ours$_{personalized}$ refer to the grasps generated when using template and personalized text description inputs, respectively. $\uparrow$ denotes higher values are better, $\downarrow$ denotes lower values are better.}
\label{tab:sota}
\end{table*}

\begin{figure*}[t]
\centering
\includegraphics[width=0.96\textwidth]{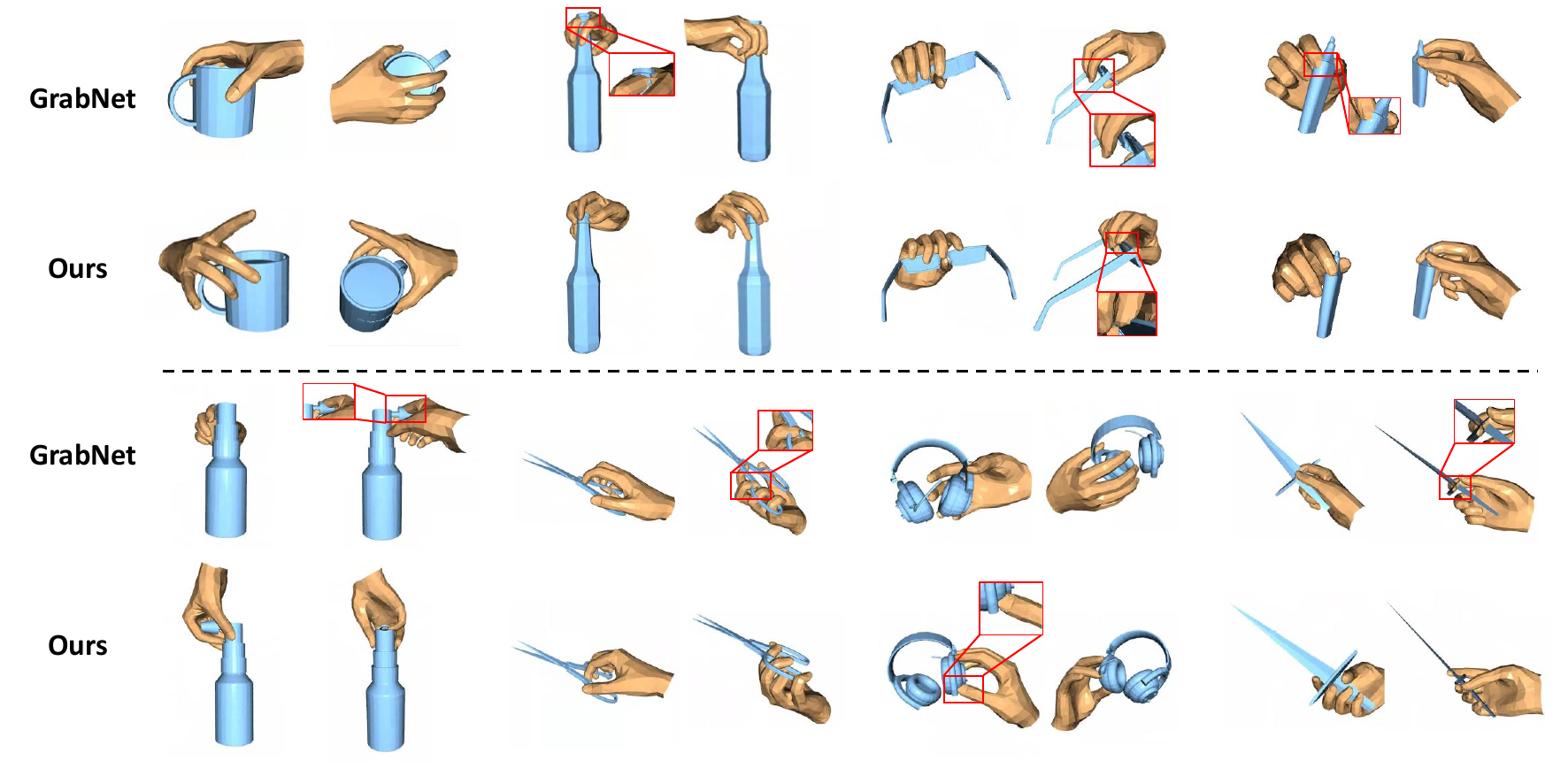} 
\caption{The qualitative results on the OakInk\cite{yang2022oakink} dataset and the AffordPose \cite{jian2023affordpose} dataset. The results demonstrated above
the dotted line are from OakInk \cite{yang2022oakink} dataset, while below are from AffordPose \cite{jian2023affordpose} dataset.}
\label{fig:sota_single}
\end{figure*}

\section{Experiments}\label{sec:4}

In this section, we demonstrate the performance of our proposed PLAN-Grasp. We first introduce our implementation details in \cref{sec:4.1}, followed by the used datasets and evaluation metrics in \cref{sec:4.2} and \cref{sec:4.3}, respectively. In \cref{sec:4.4}, we compare our method with the state-of-the-art methods and various applications that we can support. Finally, in \cref{sec:4.5}, we conduct ablation studies to verify the effectiveness of components we design.

\subsection{Implementation Details}\label{sec:4.1}

We conduct all the experiments using a single NVIDIA GeForce RTX4090 GPU with 24G memory. We sample N=2048 points sampling from the object surface as the input object points. During the training, we use the Adam optimizer \cite{kingma2014adam} with the learning rate of 1e-4 to train the denoising network LAN-GraspDiff for 1000 epochs. The training batch size is 64. Following Scenediffuser\cite{huang2023diffusion}, we set the diffusion step $T$ to 100 , which is enough for a single 3D hand pose. During the refinement stage, we utilize Adamax \cite{kingma2014adam}  to optimize the grasp vector, applying different learning rates for its components: 1e-2 for hand pose, 1e-5 for hand shape, and 1e-4 for the distance between hand and object centroids, across a total of 200 epochs.  

\subsection{Dataset}\label{sec:4.2}

\textbf{OakInk.} The OakInk\cite{yang2022oakink} is a large-scale dataset that captures hand-object interactions oriented around 5 intents: use, hold, lift-up, hand-out, and receive. It provides 1800 object models of 32 categories with their part labels and interacting hand poses. We use the shape-based subset Oak-Shape to conduct experiments, 1308 objects for training and 183 objects for evaluation. 

\textbf{AffordPose.} The AffordPose\cite{jian2023affordpose} is a large dataset of hand-object interactions with 8 affordance-driven labels such as twist, lift, and press. It comprises 641 objects from 13 categories in PartNet \cite{mo2019partnet} and PartNet-Mobility\cite{xiang2020sapien}. To evaluate the generalization ability of our method, we select 6 object categories identical to OakInk\cite{yang2022oakink}: bottle, disperser, earphone, knife, mug and scissors, and randomly chose 30 instances from each category for testing.

\begin{figure*}[t]
\centering
\includegraphics[width=0.96\textwidth]{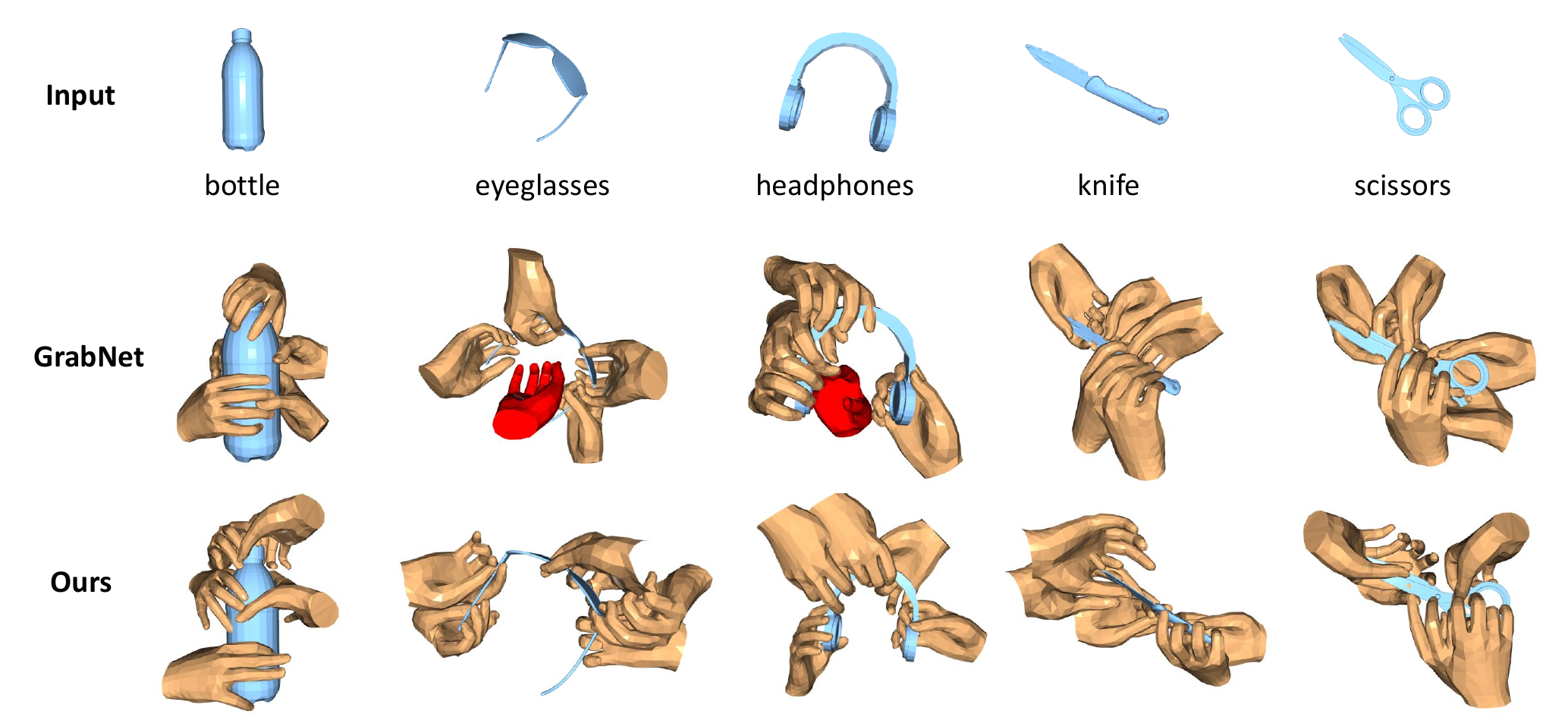} 
\caption{The qualitative results of the diverse grasps on the objects. For each object, we visualize five grasps and the red shape represents abnormal grasps.}
\label{fig:sota_multi}
\end{figure*}

\begin{figure*}[t]
\centering
\includegraphics[width=0.96\textwidth]{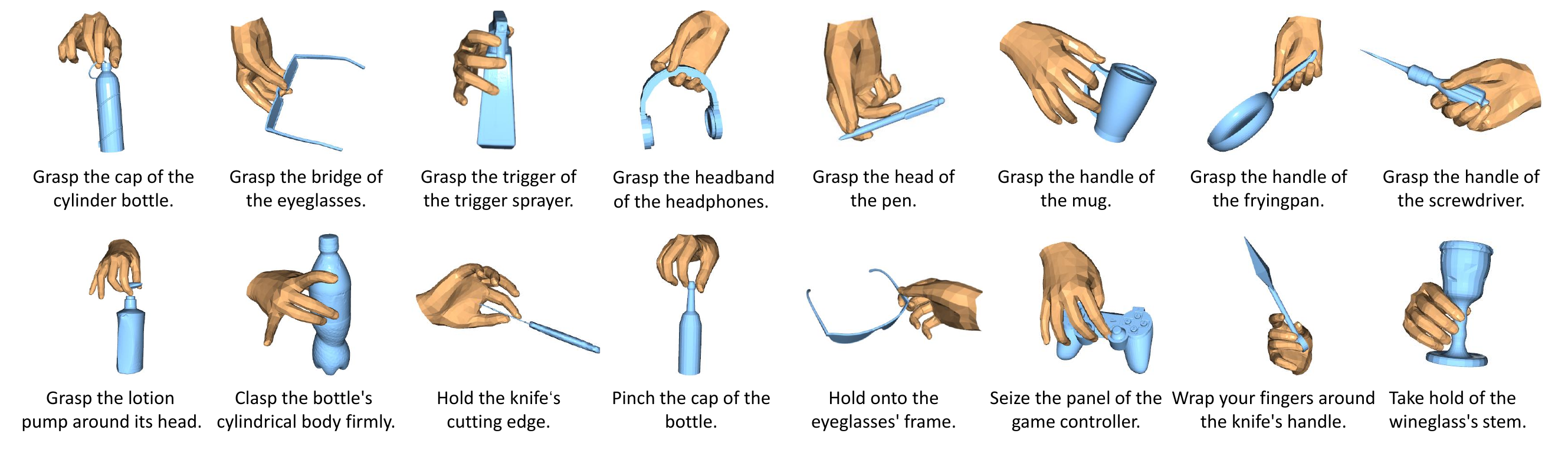} 
\caption{Visualization of the grasps generated when using different types of text inputs.  The top row displays grasps produced from  template text inputs, while the bottom row exhibits those generated from personalized text inputs.}
\label{fig:sota_language}
\end{figure*}

\subsection{Metrics}\label{sec:4.3}

A superior text-guided grasp should not only securely hold the object but also grasp the correct object part specified by text prompts. In this work, we adopt 4 metrics in total cover both grasp quality and grasp part accuracy.

\textbf{Penetration.} Following \cite{yang2021cpf,yang2022oakink,li2022contact2grasp}, we compute the Penetration Depth  and the Solid Intersection Volume between hand and object to measure the hand-object penetration. The PD is the maximum distance of all the penetrated hand vertices to their closet object surface, and the SIV is calculated by summing the volume of object voxels that are inside the hand surface.

\textbf{Simulation Displacement.} Following \cite{liu2023contactgen,hasson2019learning,yang2022oakink}, we place the object and the predicted hand into simulator\cite{coumans2015bullet}, and calculate the displacement of the object center over a period of time by applying gravity to the object.

\textbf{Diversity.} Following \cite{karunratanakul2021skeleton,liu2023contactgen}, we measure the diversity by clustering generated grasps into 20 clusters using K-means, and calculate the entropy of the cluster assignments and the average cluster size. 

\textbf{Grasp Part Accuracy.} Employing the approach introduced in \cref{sec:3.2}, we assign text template to each generated grasp and determine their accuracy by comparing with the input text descriptions. Grasp Part Accuracy is defined as the ratio of correctly identified grasps to the overall number of the generated grasps.

\begin{table*}[ht]
\centering
\begin{tabular}{m{5cm}cccccccc}
\toprule
\rule[-1ex]{0pt}{4ex} \multirow{2}{*}{\textbf{Methods}} & \multicolumn{2}{c}{\textbf{Penetration}} &  \multirow{2}{*}{\begin{tabular}[c]{@{}c@{}}\textbf{Simulation} \\ \textbf{Displacement} \\ \textbf{Mean ± Var$\downarrow$}\end{tabular}}  & \multicolumn{2}{c}{\textbf{Diversity}} &  \multirow{2}{*}{\begin{tabular}[c]{@{}c@{}}\textbf{Part} \\ \textbf{Accuracy$\uparrow$}\end{tabular}}    \\ \cmidrule{2-3} \cmidrule{5-6} \rule[-1ex]{0pt}{4ex}
 & \textbf{Depth$\downarrow$} & \textbf{Volume$\downarrow$} &  & \textbf{Entropy$\uparrow$} & \textbf{Cluster Size$\uparrow$}     \\ 
\midrule
Base.(VAE)               &         0.55 &	8.44 &	\textbf{2.48±2.56}	&2.90&	3.15	&77.38    \\ 
Base. (Diffusion)             &     \textbf{0.38}	& \textbf{2.82}	&3.00±3.04	& \textbf{2.93}	&\textbf{3.46}	&\textbf{85.25}  \\ 
 \cmidrule{1-7}
Base. w/ Add fusion   &  0.38   &   2.89   &  3.09±2.92 &  2.90  &   3.45 & 83.44  \\ 
Base. w/ Attention fusion     &   0.38	&\textbf{2.82}	&\textbf{3.00±3.04}	&\textbf{2.93}	&\textbf{3.46}	&\textbf{85.25}     \\ 
 \cmidrule{1-7}
Base. (Diffusion)     &        0.38	&2.82	&3.00±3.04	&2.93	&3.46	&85.25   \\ 
Base. + opt. w/ global   &      0.39 &	1.82&	\textbf{2.39±2.34}	&2.95&	4.18	& 83.69      \\ 
Base. + opt. w/ finger perception   &      0.38 &	1.79&	2.50±2.56&	\textbf{2.95}	&4.41	&83.74      \\ 
Base. + opt. w/ finger perception and  object part  perception    &  \centering 0.40 &	1.89	&2.49±2.51& 2.92	&\textbf{4.70}	&\textbf{87.76}      \\ 
  Base. + opt. w/ refinenet\cite{taheri2020grab}    &     \textbf{0.30}	&\textbf{1.38}&	3.11±2.81	&2.87	&2.86&	83.55  \\ 
                                                 
\bottomrule
\end{tabular}
\caption{The quantitative results of ablation study on the OakInk \cite{yang2022oakink} dataset. $\uparrow$ denotes higher values are better, $\downarrow$ denotes lower values are better.}
\label{tab:alabtion}
\end{table*}

\begin{figure*}[t]
\begin{minipage}[t]{0.48\textwidth}
\centering
\includegraphics[width=0.98\textwidth]{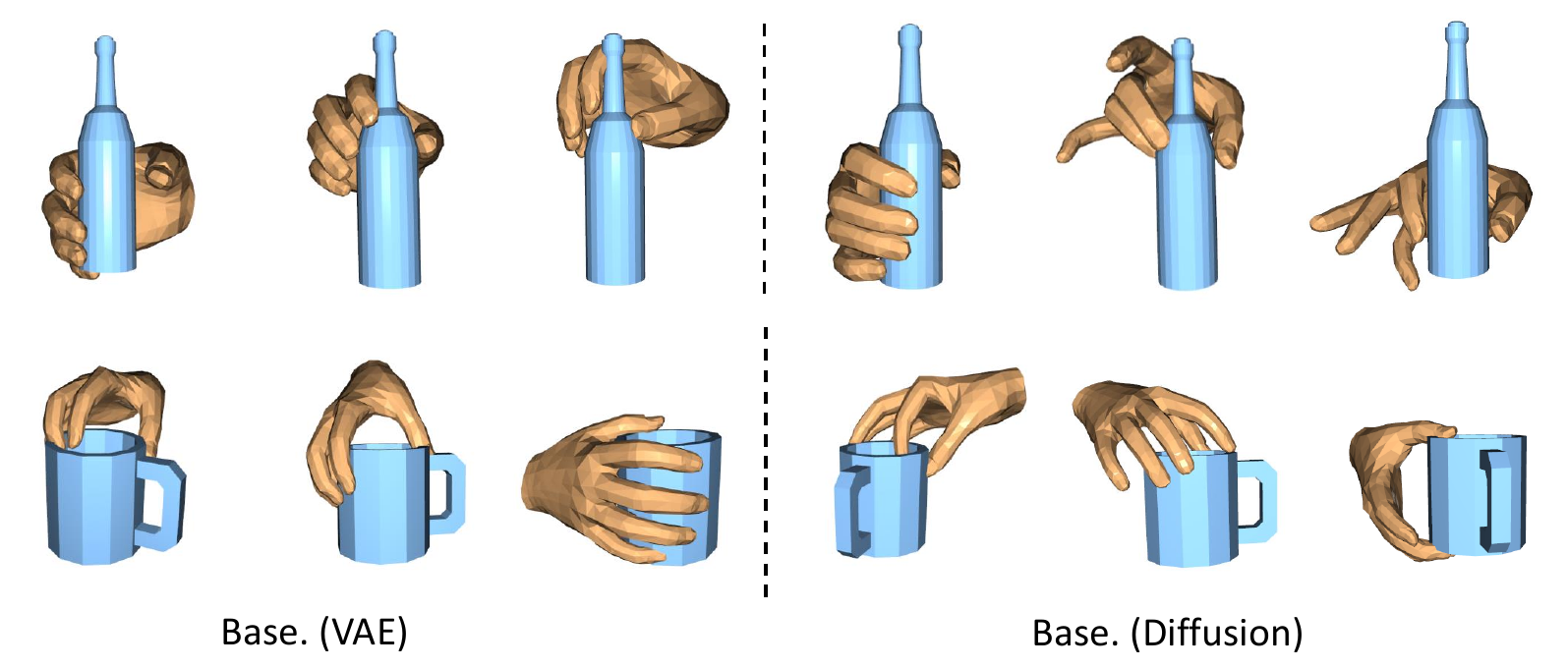} 
\caption{Comparisons of ours based on VAE(Base. (VAE)) and diffusion model (Base. (Diffusion)).}
\label{fig:ablation_diffusion}
\end{minipage}
\begin{minipage}[t]{0.48\textwidth}
\centering
\includegraphics[width=0.98\textwidth]{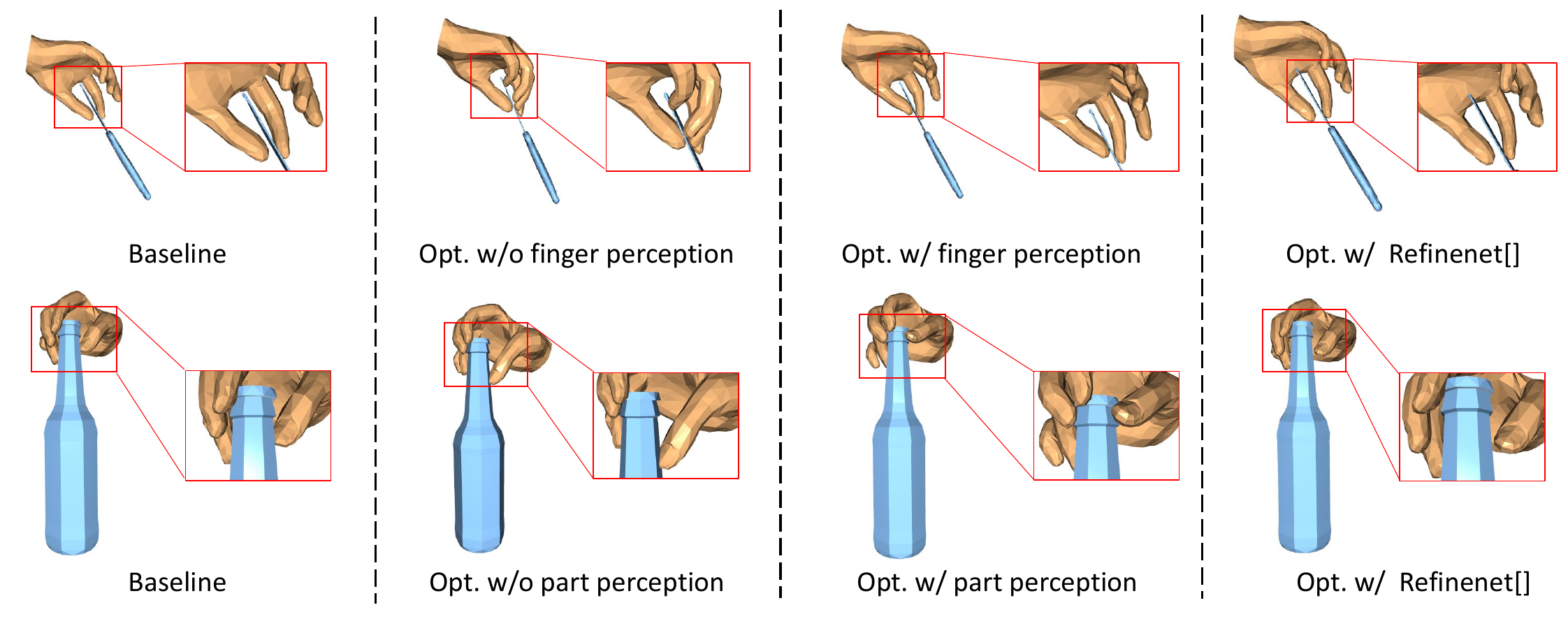} 
\caption{Comparisons of different optimization strategies. }
\label{fig:ablation_refine}
\end{minipage}
\end{figure*}

\begin{figure}
    \centering
    \includegraphics[width=0.49\textwidth]{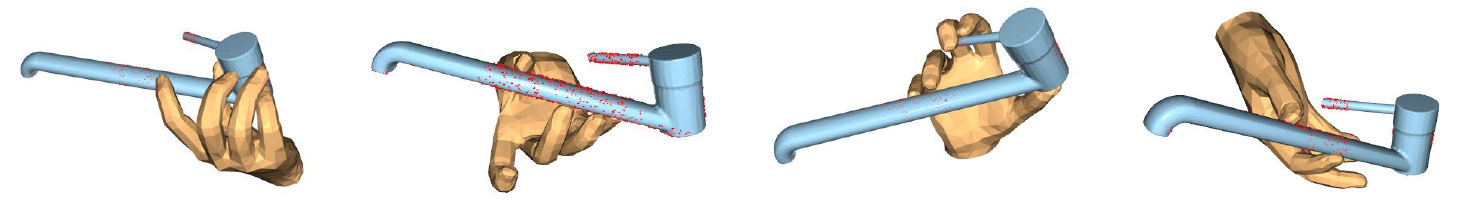}
    \caption{Visualization of the grasps generated from the text input \textit{'grasp the handle of the faucet'} given an unseen category faucet.}
    \label{fig:discussion}
\end{figure}

\subsection{Comparison with the State-Of-The-Arts}\label{sec:4.4}

To evaluate the controllability of grasp synthesis in our method, we utilize two class-level public datasets, OakInk\cite{yang2022oakink} and AffordPose\cite{jian2023affordpose}, comprising multiple categories and instances within each category. Instance-level datasets like Grab\cite{taheri2020grab} and HO3D\cite{hampali2020honnotate} are unsuitable for evaluating our method because they contain only one instance per category, and only achieve controllability on a single instance cannot verify our generalization ability. For a fair comparison, we compare the state-of-the-art method trained on OakInk\cite{yang2022oakink}: GrabNet\cite{taheri2020grab}, which is used for grasp generation in newest work \cite{yang2022oakink,jian2023affordpose}. We train it on the OakInk dataset using its officially released code and test it and our method on 183 unseen objects from the OakInk dataset and 180 out-of-domain objects from the AffordPose dataset. Following \cite{karunratanakul2021skeleton,liu2023contactgen}, we generate 20 hand grasps for each test object. Specifically for our methods, we randomly create 20 text prompts based on the parts of each test object, using these prompts and the objects as inputs to produce grasps.

We first present the quantitative comparison results on the in-domain OakInk\cite{yang2022oakink} dataset and the out-of-domain AffordPose\cite{jian2023affordpose} dataset as shown in \cref{tab:sota}. It can be seen that our method achieves the lower penetration and  simulation displacement on the OakInk dataset indicating the higher grasp quality than GrabNet\cite{taheri2020grab}. Besides, our results are close to and even outperform the ground truth in diversity that demonstrate our method achieves more diverse and natural grasps. Experimental results on AffordPose\cite{jian2023affordpose} Dataset demonstrate that our method achieves the comparable generalization ability, with the lower simulation displacement, higher diversity, and comparable penetration. More importantly, we achieve the grasp controllability by text prompts of object grasping parts, a capability not present in GrabNet \cite{taheri2020grab} and get grasp part accuracy of 87.76\% with template text and 82.32\% with personalized text on OakInk dataset\cite{yang2022oakink} as shown in \cref{tab:sota}.

Furthermore, to evaluate the performance qualitatively, we visualize the generated hand grasps for both in-domain and out-of-domain objects by GrabNet \cite{taheri2020grab} and our method. As shown in \cref{fig:sota_single}, it can be seen that both  methods can generate  plausible hand grasps for object part with sample shapes such as the body of mug and the cap of earphones. But for specific-part grasp such as the cap of bottle, we can observe clearly from the red boxes that the grasps our method generated has the smaller penetration and more natural contact. These results demonstrate our method’s capability to generate physically plausible and stable grasp. Moreover, we visualize the multi-grasps for each object in \cref{fig:sota_multi}.  In contrast to the predominantly closed hand poses with five fingers generated by GrabNet\cite{taheri2020grab},  ours is better suitable for the specific shape of the object, as demonstrated \cref{fig:sota_multi} with the eyeglasses and knife.

To show our grasp controllability, we visualize the results of grasp synthesis guided by text prompts of object grasping parts. which consist of template  and personalized text descriptions. As shown in \cref{fig:sota_language}, our method not only generates hand poses that grasp the objects of different categories in a manner consistent with human habits but also directly produces grasps in a text-controlled manner. This level of controllability enables precise object part grasping for subsequent tasks. 

\subsection{Ablation Study}\label{sec:4.5}

\textbf{VAE vs Diffusion.} To fairly evaluate the effectiveness of the  diffusion model we employed, we construct a variant of our method by replace the diffusion model with VAE for part-level grasp sythesis. Both this model and ours remove the optimization process. The results on the OakInk \cite{yang2022oakink} dataset are shown in the first two rows of \cref{tab:alabtion} and \cref{fig:ablation_diffusion}. From the experimental results we can see diffusion model achieves the higher grasp quality with lower penetration and higher grasp part accuracy. More importantly, it can be seen obviously from \cref{fig:ablation_diffusion}  that our diffusion model can generate more diverse hand pose for mug and bottle than VAE.

\textbf{Multi-Modal Attention.} We evaluate the effectiveness of the Multi-Modal Attention which is designed to fuse the text feature and the object point feature. Specifically, we compare this module with feature addition. As shown in the 3rd and the 4th of \cref{tab:alabtion}, the multi-modal attention outperforms feature addition across all metrics, especially in the grasp part accuracy.

\textbf{Optimiziation.}  We evaluate the effectiveness of optimization based on finger perception and text-guided object part perception, and the results are shown in \cref{tab:alabtion} and \cref{fig:ablation_refine}. Note that here we use Baseline (Base.) to represent our method without any optimization. It can be seen the grasp quality achieves a significant improvement by adding the global optimization which leads all-fingers toward object for Baseline. Specifically, the penetration volume and the simulation displacement have decreased by 35.46\%, and 20.33\% respectively, while diversity has improving by 20.81\%. However, it directs all fingers towards the object’s nearest point, limiting the diversity  and leading to inaccuracies in the contact part, as shown in \cref{fig:ablation_refine}. In contrast, our optimization, grounded in finger perception, fine-tunes only the fingers involved in grasping, while the rest maintain a natural state. This approach enables us to maintain grasp quality while achieving greater diversity. Furthermore, as shown in \cref{fig:ablation_refine}, the optimization based on text-guided object part perception focuses on directly the hand towards the part described by text description, enabling us to achieve a higher grasp part accuracy. 

In addition, we compare our optimization with the RefineNet used by GrabNet\cite{taheri2020grab}. It is trained to denoise on the dataset built by adding random noise into ground-truth hand-object interaction. As the \cref{fig:ablation_refine} illustrated, grasps optimized by RefineNet still do not fully contact the blade. In contrast, our method, optimized for specific situations, performs better in detail. And the quantitative results in \cref{tab:alabtion} demonstrate our methods achieve better balance between penetration and simulation displacement.

\section{Discussion}\label{sec:5}

The part control ability of our method can be easily transferred among the objects in the seen categories. However, due to the limited categories in the training dataset, when faced with the objects of new categories that have never been seen in the training dataset, although reasonable grasp can be generated, the contact parts can not be identified because of the lack of understanding of the new categories. As shown in \cref{fig:discussion}, our method can produce a reasonable grasp for the faucet but cannot accurately grasp the faucet’s handle. Therefore, it is necessary to train on a grasp dataset with more categories, but such a dataset is currently not available. In addition, with the help of the Large Language Models\cite{brown2020language}, we can achieve task-level static grasp synthesis, such as grasp the handle of the knife rather than the blade when cutting fruit. However, correctly grasping the part of an object is the first step to complete the task and the ability to dynamically manipulate objects is also the key for task. We will do more exploration in the future.

\section{Conclusion}\label{sec:6}
In this work, we introduce Text2Grasp, a grasp synthesis method guided by text prompts of object grasping parts. It begins with a text-guided diffusion model, termed TextGraspDiff, which is responsible for generating an initial, coarse grasp pose. This is subsequently refined through a hand-object contact optimization process. This method ensures that the generated grasps are not only physically plausible and diverse but also precisely aimed at specific object parts described by text prompts. Furthermore, our method also supports grasp synthesis guided by personalized text  and task-level text descriptions by LLM without extra manual annotations. Extensive experiments conducted  on two public datasets demonstrate our methods achieves not only the comparable performance in grasp quality but also precise part-level grasp control.

{\small
\bibliographystyle{ieee_fullname}
\bibliography{egbib}
}

\clearpage
\appendix
\setcounter{figure}{0}
\renewcommand{\thefigure}{A\arabic{figure}}

\onecolumn \section*{\centering Appendix} 

\begin{figure*}[ht]
    \centering
    \includegraphics[width=0.9\textwidth]{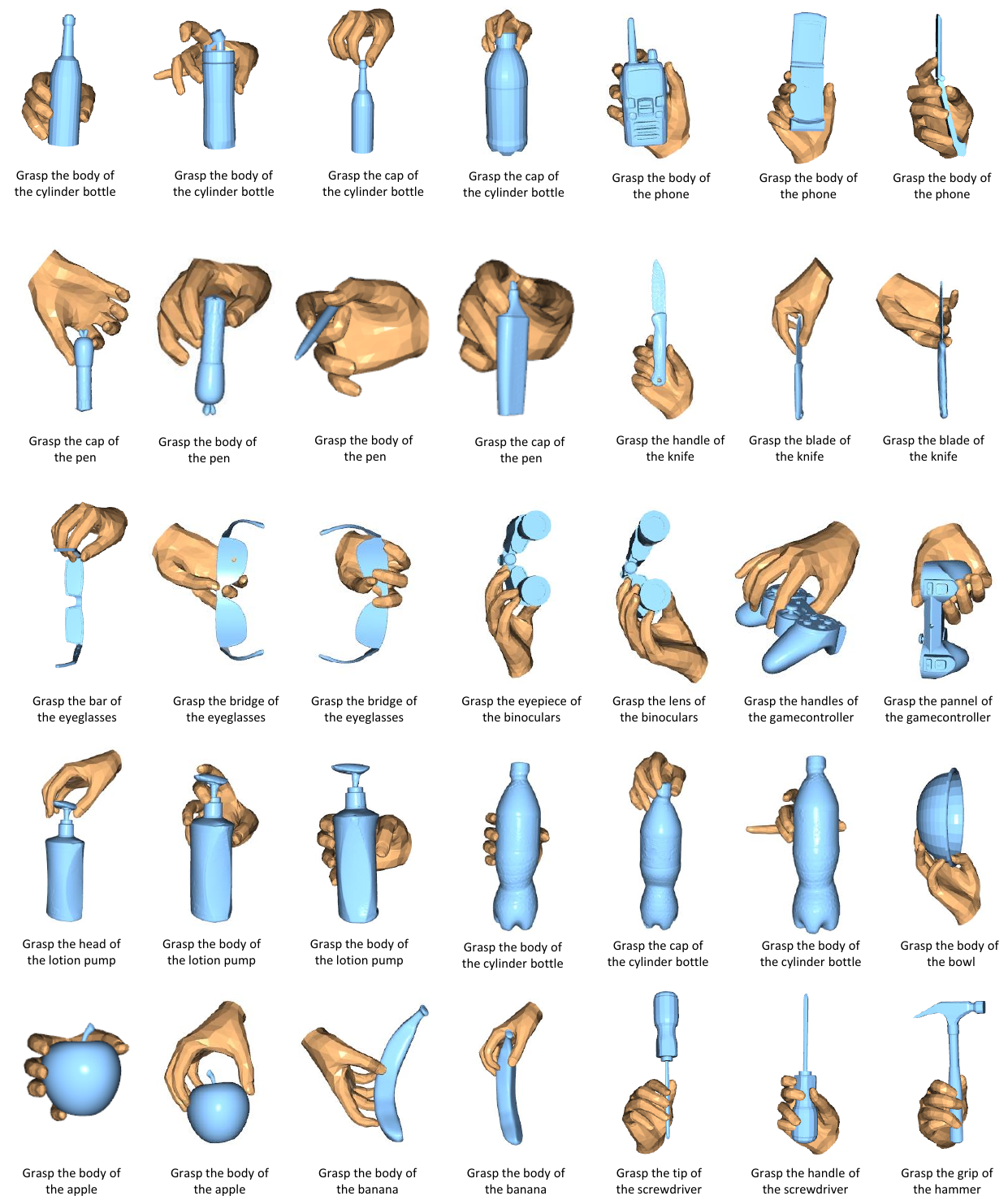}
    \caption{Visualization of the grasps generated when using template text inputs.}
    \label{fig:appendix_template}
\end{figure*}

\begin{figure*}[ht]
    \centering
    \includegraphics[width=0.9\textwidth]{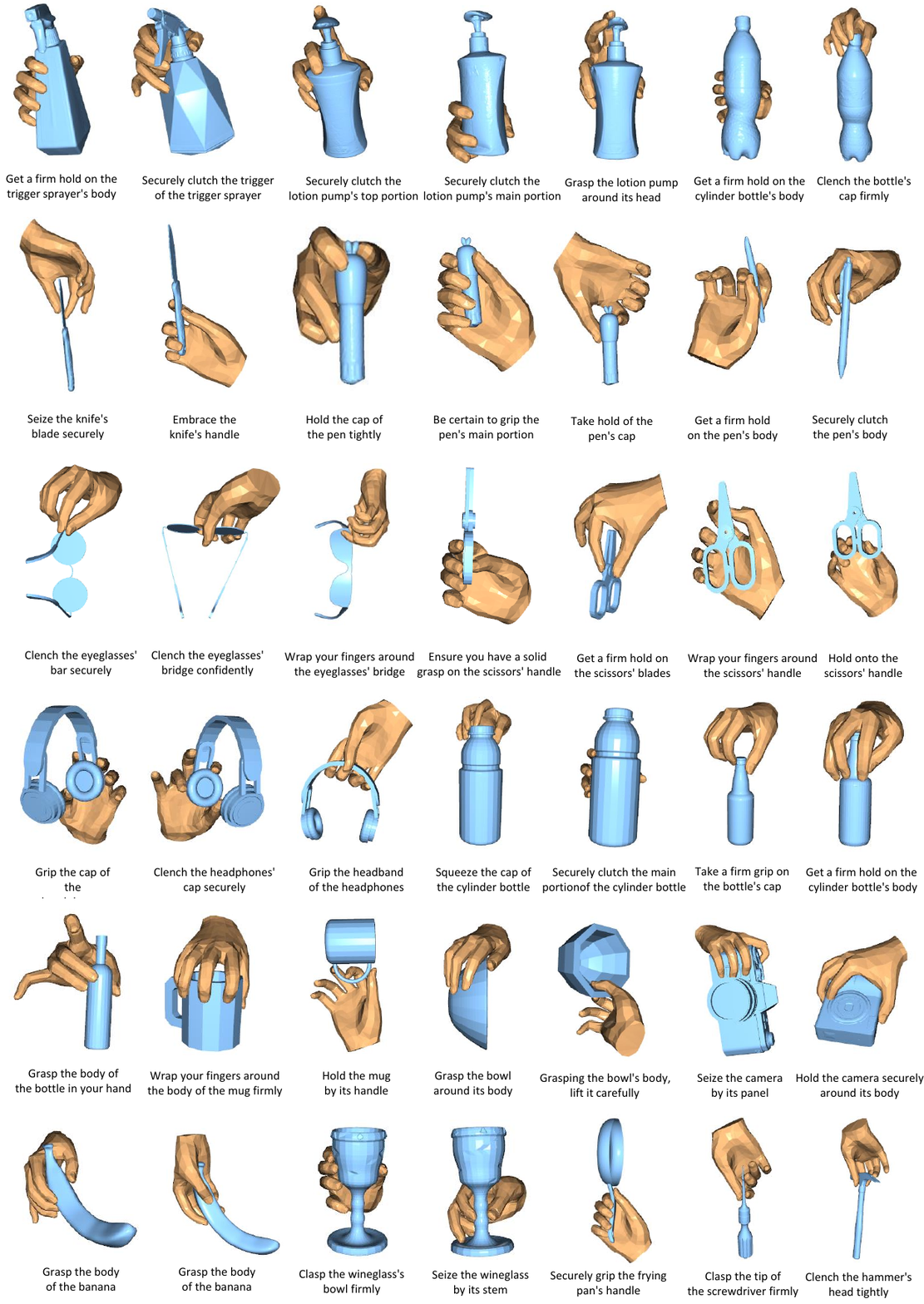}
    \caption{Visualization of the grasps generated when using personalized text inputs.}
    \label{fig:appendix_personalized}
\end{figure*}

\end{document}